%% file: main.tex
\documentclass[runningheads]{llncs}
\usepackage{graphicx}
\usepackage{xcolor}
\usepackage{subcaption}
\usepackage{caption}
\usepackage{hyperref}
\begin{document}
%
\title{No-brainer: Morphological Computation driven Adaptive Behavior in Soft Robots}
\titlerunning{No-brainer: Morphology-driven Adaptive Behavior}
%
\author{
Alican Mertan\inst{1}\orcidID{0000-0002-5947-8397} 
\and
Nick Cheney\inst{1}\orcidID{0000-0002-7140-2213}
}
\authorrunning{A. Mertan and N. Cheney}
%
\institute{University of Vermont, Burlington VT 05401, USA\\
\email{\{alican.mertan,ncheney\}@uvm.edu}}
\maketitle              
\begin{abstract}
 It is prevalent in contemporary AI and robotics to separately postulate a brain modeled by neural networks and employ it to learn intelligent and adaptive behavior. 
 While this method has worked very well for many types of tasks, it isn't the only type of intelligence that exists in nature. 
 In this work, we study the ways in which intelligent behavior can be created without a separate and explicit brain for robot control, but rather solely as a result of the computation occurring within the physical body of a robot. Specifically, we show that adaptive and complex behavior can be created in voxel-based virtual soft robots by using simple reactive materials that actively change the shape of the robot, and thus its behavior, under different environmental cues. 
 We demonstrate a proof of concept for the idea of closed-loop morphological computation, and show that in our implementation, it enables behavior mimicking logic gates, enabling us to demonstrate how such behaviors may be combined to build up more complex collective behaviors.

\keywords{Soft robotics  \and Adaptive behavior}
\end{abstract}
\section{Introduction and Background} \input{sections/intro}


\section{Methodology} \input{sections/methods}

\section{Responding to Binary Stimuli} \input{sections/binary_stim}

\section{Making a Swarm for More Complex Behavior} \input{sections/swarm}

\section{Discussion} \input{sections/discussion}

\section{Conclusion} \input{sections/conclusion}

\subsubsection{\textbf{Acknowledgements}} This material is based upon work supported by the National Science Foundation under Grants No. 2008413 and 2239691.
Computations were performed on the Vermont Advanced Computing Core supported in part by NSF Award No. 1827314.

\bibliographystyle{splncs04}
\bibliography{all}

\end{document}

%% file: sections/intro.tex
%

Recent advances in artificial intelligence and machine learning have benefited greatly from the rise of modern deep learning systems, ultimately aimed at artificial general intelligence~\cite{lecun2015deep}.  The coming-of-age of these artificial neural network systems includes a long history of bio-inspiration, dating back to Mcculloch and Pitts~\cite{mcculloch1943logical}.  Yet the processes behind biological intelligence reach far beyond systems and processes confined to the brain of living organisms.  

Our bias toward attributing intelligent behavior to the mind is far from new.  Descartes' mind-body-dualism dates back to the 1600s~\cite{descartes_discourse_1993}.  
However, Bongard warns that ``thinking about thinking is misleading''~\cite{bongard2022er}. Moravec's paradox, the observation that reasoning takes less computational resources than sensory-motor skills (contrary to the expectations of experts), is an example of how intuition could fail in introspection on thinking~\cite{moravec1988mind}. 
Examples of complex, perhaps intelligent-seeming, behavior stemming from non-neural processes are abound in both engineered and natural systems.  Perhaps the most notable example of a system built to mimic a complex behavior in simple and purely mechanical form is the passive walking robot, which gracefully walks down an incline plane simply due to the well-tuned structure of its joints and limbs interacting with the physics of its environment~\cite{mcgeer1990passive}.  In natural systems, the growth and shape change towards stimuli in plants serve as an obvious example of non-neural behavior~\cite{karban2008plant,trewavas2003aspects}.  

More generally, these phenomena -- of the body's complex and ``intelligent'' behavior, with or without a brain -- are widely studied within the fields of embodied cognition~\cite{wilson2002six} and morphological computation~\cite{pfeifer2009morphological}.  
The literature argues that the brain is not the only source of adaptive and complex behavior, and that the body ``shapes the way we think''~\cite{pfeifer2006body}. Even for humans, who naturally consider their brains as the source of their intelligent behavior, there are strong indications that the body plays an important role in intelligence, for example, natural language, something we consider one of the fundamental products of our intelligence, is deeply connected to our bodies~\cite{lakoff2008metaphors,pulvermuller2010active}. 

Prior works have demonstrated the concept of morphological computation in physical bodies, designing or evolving a body plan which executes a given behavior with as little neural or control intelligence as possible~\cite{auerbach2010evolving,cheney2015evolving,cheney2013unshackling,corucci_material_2016,hauser2011towards,joachimczak2016artificial,paul2006morphological}, but these body plans tend to focus on finding an effective morphology for a single desired behavior.
Conversely, very simple adaptive behaviors in neural circuits demonstrate minimally cognitive systems that adapt the behavior of a fixed body plan to environmental stimuli via extraordinary simple neural circuitry~\cite{beer1996toward,braitenberg1986vehicles}.  
Bridging the gap between the two, prior work on brain-body co-optimization attempt to simultaneously evolve simple controllers and evolving robot bodies~\cite{hornby2001body,bongard2003evolving,joachimczak2012co,lehman2011evolving,sims2023evolving} or develop approaches to rapidly adapt a controller to an ever-evolving body plan~\cite{cheney_scalable_2018,mertan2023modular,mertan2024investigating}.
But, to the best of our knowledge, prior work thus far has failed to demonstrate an example of adaptive real-time behavior stemming entirely from morphological computation (i.e. with no controller or neural circuitry).   

Inspired by the diverse forms of complex behaviors in nature, here we study ways in which adaptive behavior can be created in an artificial agent without a separately postulated brain. Particularly, we evolve voxel-based soft robots~\cite{bhatia2021evolution} 
for the task of locomotion to study how morphology could be a source of adaptive behavior. 
Four stimuli-responsive materials change shape in response to environmental cues, resulting in a gross morphological change to the robot.  Robots are optimized via an evolutionary algorithm to produce body-plans that adaptively change direction on-the-fly in response to combinations of these environmentally-driven morphology changes (Fig.~\ref{fig:sensory_voxels}), resulting in closed-loop robot behaviors with no controller or neural circuitry (Figs.~\ref{fig:champions-behavior-fixed},~\ref{fig:behavior_comparison},~\ref{fig:sample-behavior}), that can be combined to create complex morphological logic gates (Fig.~\ref{fig:swarm-system}).  

%% file: sections/methods.tex
\textbf{Simulation:} We run our experiments on the EvoGym simulator~\cite{bhatia2021evolution}. It is a 2D soft robot simulator where robots are represented by voxels that can actuate by changing their areas ($a \in [0.6 \times r, 1.6 \times r]$ where $r$ is the resting area). Designing robots' bodies means placing voxels with varying materials in a grid layout and specifying connections with their neighbors. In our experiments, we assume all neighboring voxels are connected to simplify the design space.

\begin{figure}[t]
    \centering
    \begin{subfigure}{0.25\textwidth}
        \centering \captionsetup{width=.9\linewidth}
        \includegraphics[width=0.45\textwidth]{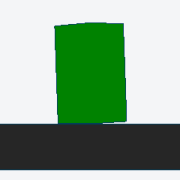}%
        \includegraphics[width=0.45\textwidth]{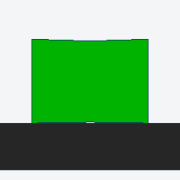}%
        \caption{Sensor type 1 (stimulus 1)}
    \end{subfigure}%
    \begin{subfigure}{0.25\textwidth}
        \centering \captionsetup{width=.9\linewidth}
        \includegraphics[width=0.45\textwidth]{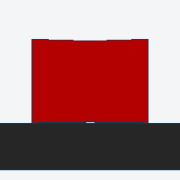}%
        \includegraphics[width=0.45\textwidth]{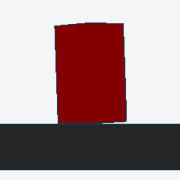}%
        \caption{Sensor type 2 (stimulus 1)}
    \end{subfigure}%
    \begin{subfigure}{0.25\textwidth}%
    \centering \captionsetup{width=.9\linewidth}
        \includegraphics[width=0.45\textwidth]{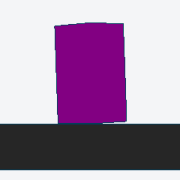}%
        \includegraphics[width=0.45\textwidth]{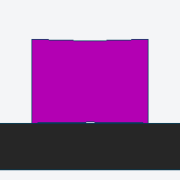}%
        \caption{Sensor type 3 (stimulus 2)}
    \end{subfigure}%
    \begin{subfigure}{0.25\textwidth}
    \centering \captionsetup{width=.9\linewidth}
        \includegraphics[width=0.45\textwidth]{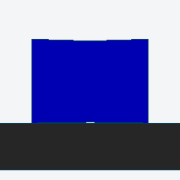}%
        \includegraphics[width=0.45\textwidth]{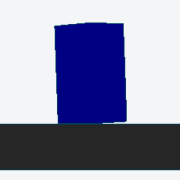}%
        \caption{Sensor type 4 (stimulus 2)}
    \end{subfigure}%
    \\
    \begin{subfigure}{\textwidth}
    \centering
    \includegraphics[width=0.250\textwidth]{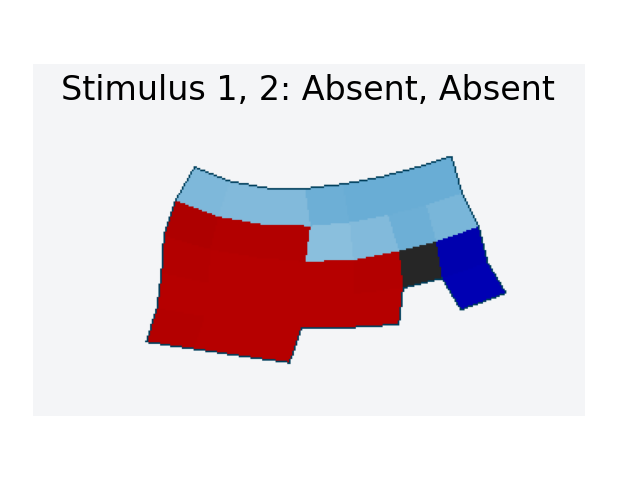}%
    \includegraphics[width=0.250\textwidth]{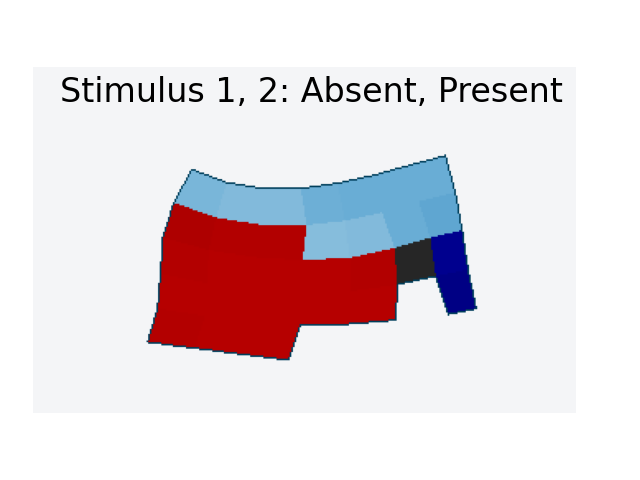}%
    \includegraphics[width=0.250\textwidth]{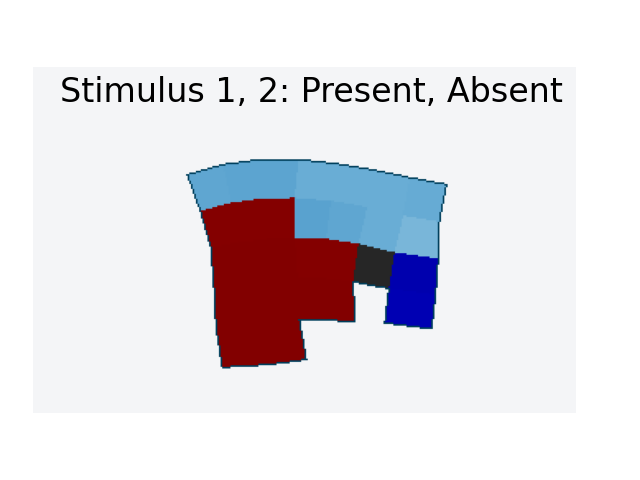}%
    \includegraphics[width=0.250\textwidth]{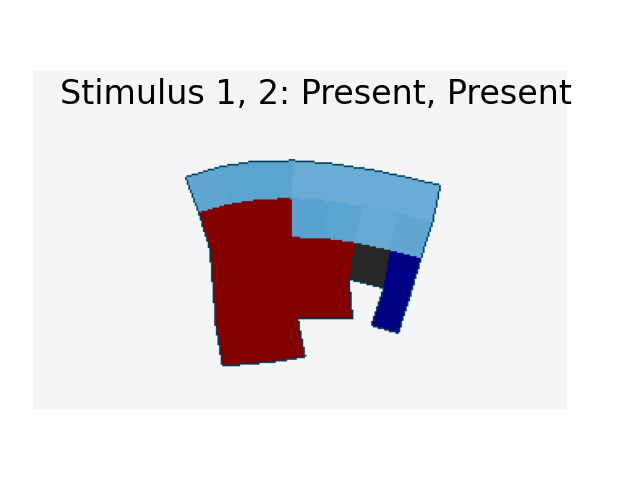}%
    \\
    \includegraphics[width=0.250\textwidth]{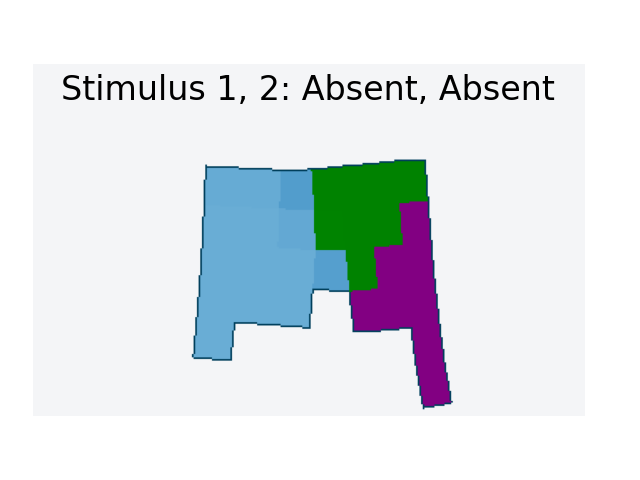}%
    \includegraphics[width=0.250\textwidth]{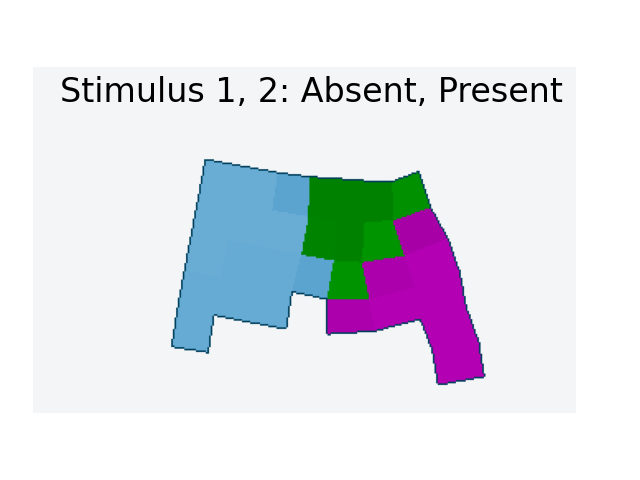}%
    \includegraphics[width=0.250\textwidth]{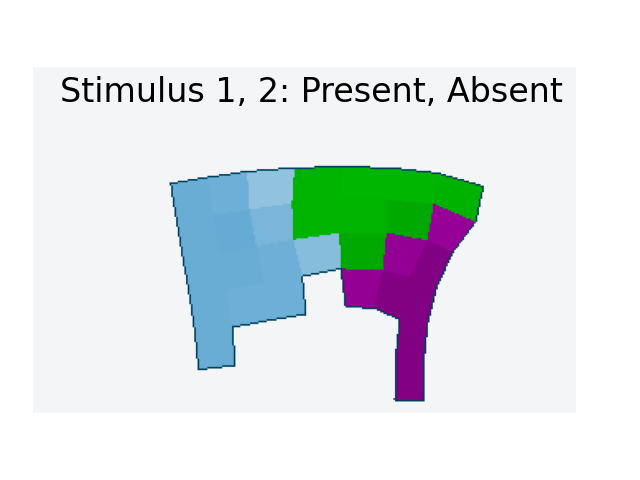}%
    \includegraphics[width=0.250\textwidth]{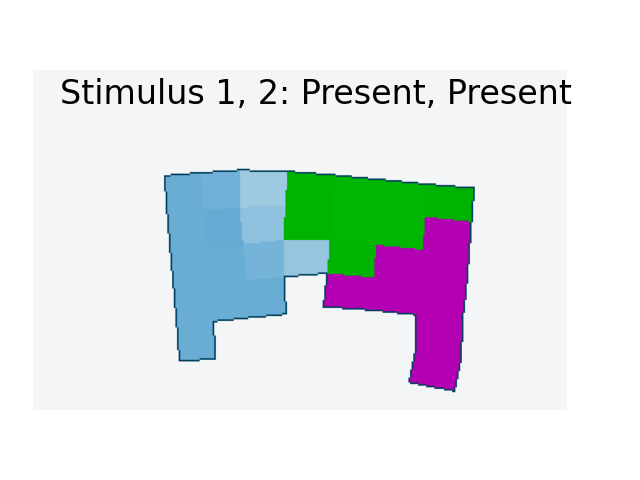}%
    \caption{Example morphologies under different stimuli patterns. The behavior of sensory voxels molds the morphology of the robot, potentially changing its behavior.}
    \end{subfigure}
    \caption{Images from the simulator environment~\cite{bhatia2021evolution} that shows the behavior of sensory voxels. Voxels in (a) and (b) respond to stimulus 1 and voxels in (c) and (d) respond to stimulus 2. Voxels in (a) and (c) shrink to their minimum width in response to the absence of their respective stimulus (left), and expand to their maximum width in response to the presence of their respective stimulus (right). Voxels in (b) and (d) respond the opposite way. These materials enable robots (e) that actively undergo shape change to produce new behaviors according to cues from some sensory stimuli.
    }
    \label{fig:sensory_voxels}
\end{figure}

\textbf{Environment:} We use EvoGym's Python API to define a custom environment with two stimuli. These stimuli work in a binary fashion -- they are either present or absent in the environment. For our robots to sense these stimuli and adapt their behavior, we hand-design "sensory voxels". 
We consider two environmental stimuli, each with one voxel type that expands in response to that stimuli and one that contracts in the presence of that stimuli -- for a total of 4 sensory voxel types (Fig.~\ref{fig:sensory_voxels}; green, red, magenta, blue).
 We also hand-designed two active voxels (orange and teal) that horizontally expand and contract to provide energy for locomotion following a sinusoidal signal with a 180\textdegree\ phase offset, similar to the designs in~\cite{cheney2013unshackling}, but do not provide any neural or informational connectivity between them to organize a coordinated controller by means other than physical forces traveling through the robot's body. 
Lastly, we provide evolution with two passive voxels, soft (gray) and rigid (black).

\textbf{Evolutionary algorithms}~\cite{eiben2015introduction} are the most common approach to optimize robot body plans in the above-cited prior work, as they work by evaluating the empiric and holistic phenotype/behavioral effect of random mutations of a robot's artificial genome, selecting the most fit robots to survive and reproduce -- and thus do not require a differentiable model of the robot's morphology for credit assignment of the robot's fitness to individual voxels.  Here we employ the Map-Elites algorithm~\cite{mouret_illuminating_2015} to evolve an archive of diverse robots.
Specifically, the archive consists of seven feature dimensions. Three of these dimensions are based on morphological features -- the number of active voxels, sensory voxels, and total number of voxels. Following the common
practice~\cite{cheney2013unshackling},
we limit the morphology space with a $H\times W$ bounding box. Limiting the total number of voxels in this way allows us to discretize the three morphological dimensions of the archive into four bins each. The next four dimensions are based on the behavior of the robot, the direction of locomotion in the x-axis, under different stimuli patterns. The two binary stimuli result in four patterns, and a robot can move in either the positive x or the negative x direction, resulting in 16 possible behaviors in total. By utilizing QD algorithms and defining the dimensions of the archive in the described way, we allow evolution to search for all possible behaviors with varying morphological structures in a single run, and keep the best-performing robot for each unique combination of morphological-behavioral traits, via the MAP-Elites algorithm.

\textbf{Evaluation and selection:} Individuals are evaluated under 5 simulated conditions. First, they are evaluated for 40 actuation cycles (i.e. 40 periods of the sinusoidal actuation signal) where at each 10\textsuperscript{th} actuation cycle the environmental stimuli change as follows: (Stimulus 1, Stimulus 2) = \{(Absent, Absent), (Absent, Present), (Present, Absent), (Present, Present)\}. In the next 4 simulations, individuals are simulated under each of these 4 fixed stimuli patterns for 10 actuation cycles each. We use multi-objective selection to select robots that maximize total displacement in the 40-cycle variable stimuli simulation and maximize the minimum displacement across the 4 single stimuli 10-cycle simulations. Inspired from~\cite{mace2023quality}, an offspring replaces an individual from the archive, only if it dominates the individual, i.e. is better than the individual in both fitness values.  
In order to also select for robots with dynamically stable gaits that are not overly dependent on initial conditions, we disregard all robots that move in different directions for the same stimuli pattern in combined vs. single stimuli simulations.

\textbf{Genome:} Similar 
to~\cite{cheney2013unshackling,corucci_material_2016},
robot morphologies are represented by compositional pattern-producing networks (CPPNs)~\cite{stanley_compositional_2007}. CPPN is a directed acyclic graph, where each node represents a math function. Here we use the
sine, absolute value, square, and square-root as activation functions within each node.  
The CPPN takes the normalized coordinates of a voxel as well as its distance to the center of the bounding box as an input and outputs a scalar value for each possible voxel type, including the empty voxel. The type indicated by the max output value is assigned to the queried location. During evolution, offspring are created through mutation by the addition/removal of nodes/links, change of activation functions at nodes, and change of weights in edges.

\textbf{Parameters:} We experiment with three different bounding boxes, $(H,W) \in \{(5,5), (7,7), (10,10)\}$, and repeat each experiment 10 times. Evolution was run for 3000 generations where each generation 16 individuals were chosen as parents from the archive uniformly randomly to create 16 offspring through mutation. Code to repeat our experiments is 
available at \href{https://github.com/mertan-a/no-brainer}{github.com/mertan-a/no-brainer}. 

%% file: sections/binary_stim.tex
\begin{figure}[t]
    \centering
    \includegraphics[width=\textwidth]{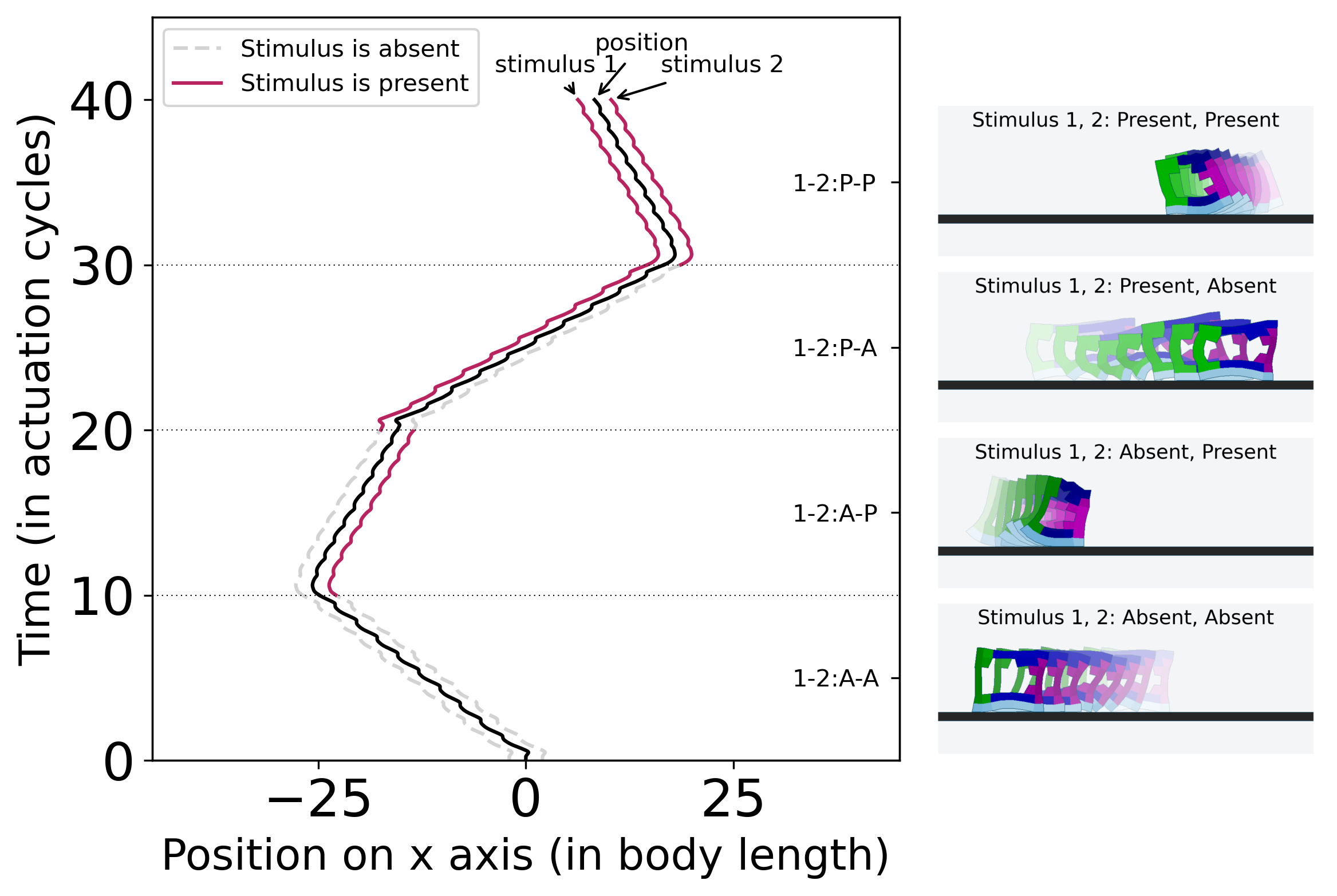}
    \caption{The spacetime diagram of a run champion (left) and its gait under different stimuli patterns as snapshots from the simulation (right). In the spacetime diagram, the solid black line in the middle keeps track of the robot's center of mass over time. The two accompanying lines show the current stimuli pattern over time. The robot moves toward the positive x-direction when one stimulus is present, otherwise, it moves in the opposite direction. The different gait behaviors are achieved by the morphological changes caused by sensory voxels -- an example of morphological computation and how the body could be the source of adaptive behavior. See more robots in action 
    at \href{https://github.com/mertan-a/no-brainer/tree/master/robots-in-action}{our repository.}
    }
    \label{fig:sample-behavior}
    \vspace{-2em}
\end{figure}

Firstly we investigate whether purely morphological changes can result in behavior that adapts to environmental stimuli. Over the 4 stimuli patterns presented to the robot, our MAP-Elites archive collects different combinations of behavioral patterns that different morphologies produce (labeled for whether a robot moves (L)eft or (R)ight for each of the 4 stimuli). 
Fig.~\ref{fig:sample-behavior} visualizes the behavior of a robot exhibiting LRRL behavior as a spacetime diagram to investigate how the morphology gives rise to adaptive behavior.
The solid line tracks the position of the robot's center of mass during simulation. The two lines accompanying the solid line display the stimuli pattern at each time step. The dotted horizontal line shows the moments when the stimuli pattern changes. 
The robot's morphology and its gait under different stimuli can be seen on the right. The robot moves to the right only if one of the two stimuli is present (the middle two settings). In the case where both stimuli are present or absent (the first and last setting), the robot moves to the left. This behavior is achieved with the help of sensory voxels -- the change in the sensory voxels alters the robot's shape and thus changes its gait. The robot exhibits an adaptive behavior without the help of a dedicated brain. Reactive materials are enough to unlock morphological computation.

\begin{figure}[t]
    \centering
    \includegraphics[width=\textwidth]{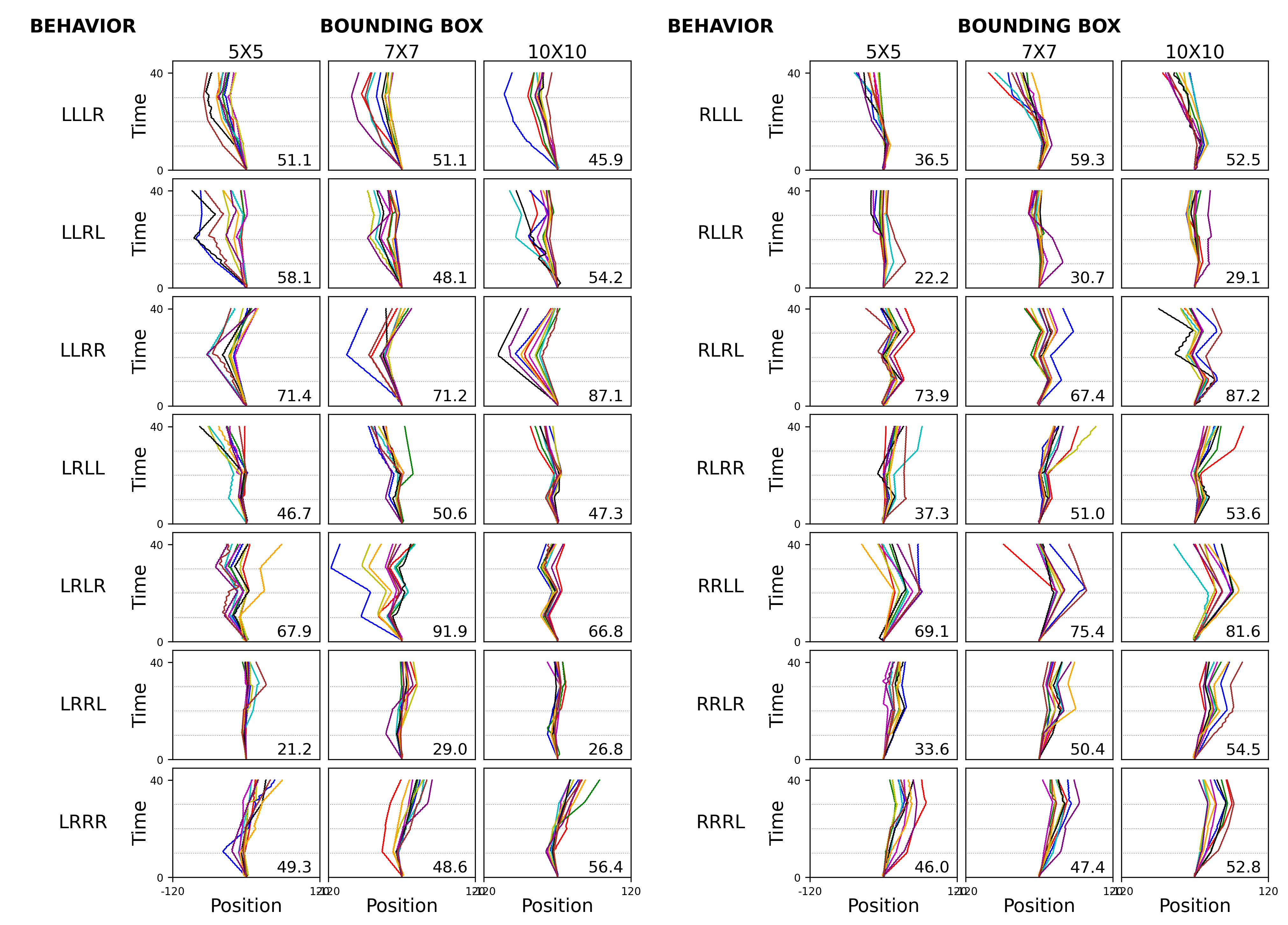}
    \caption{The spacetime diagrams of all run champions from independent runs, plotted together. We omit the behaviors where the robot doesn't respond to any stimulus (LLLL and RRRR behaviors). Time is normalized to reflect actuation cycles and position is normalized to reflect the robot's body length. The numbers in the bottom right corner show the average distance traveled by run champions. }
    \label{fig:champions-behavior-fixed}
    \vspace{-1em}
\end{figure}

Fig.~\ref{fig:champions-behavior-fixed} displays spacetime diagrams of all run champions successfully exhibiting different adaptive behaviors. Each run champion is drawn in a different color and the average normalized space traversed by run champions in each plot can be seen in the bottom right corner. We see that given adequate environmentally-sensitive building blocks to work with, evolution can find morphologies that exhibit various adaptive behaviors -- responding to stimuli in different ways via shape change alone (without a learned and/or stimuli-aware controller).

\begin{figure}[t!]
    \centering
    \begin{subfigure}{\textwidth}
        \centering 
        \includegraphics[width=\textwidth]{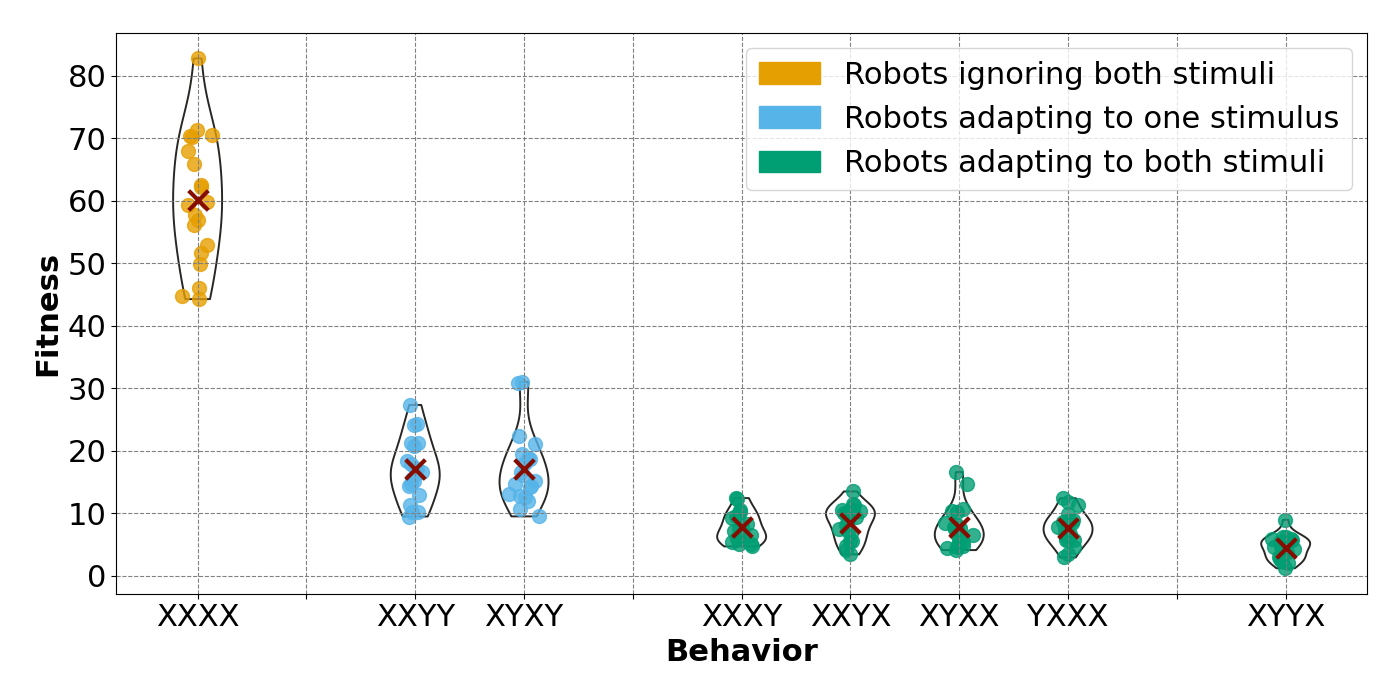}
    \end{subfigure}
    \caption{Performances of run champions from 10 independent runs with a $10\times10$ bounding box. Symmetric robot behaviors are grouped by the pattern of their movement direction across stimuli (e.g. LRLR and RLRL are both represented as XYXY). Colors represent the number of stimuli a robot's morphology adapts to. Each data point is plotted and mean values are marked with an 'x'. Behaviors are separated with empty columns based on statistically significant performance differences (at $P < 0.001$) -- each distribution in a group is statistically significantly better compared to each other distribution belonging to any group to their right. Bounding boxes of size $5\times5$ and $7\times7$ are qualitatively consistent.}
    \label{fig:behavior_comparison}
    \vspace{-1em}
\end{figure}

To analyze the effect of different behavior patterns, we group the run champions from independent runs based on the directions of their movement trajectories across the 4 stimuli. For this analysis, we group behaviors that are symmetric in the x-axis (e.g. LLLL and RRRR are grouped as XXXX, LLRR and RRLL are grouped as XXYY, etc.) as the challenge of achieving symmetric behaviors is equal. As can be seen in Fig.~\ref{fig:behavior_comparison}, robots ignoring both stimuli and moving in the same direction throughout (behavior XXXX; mustard color) are able to locomote further compared to robots that exhibit a sensory-dependent behavior. Similarly, robots that respond only to a single stimulus (behavior patterns XXYY and XYXY; light blue color) perform better compared to robots adapting their behavior following both stimuli (green). As expected, there is no statistically significant difference between following the first stimulus (XXYY) or the second (XYXY). Adapting behavior only when one of the two stimuli is present but not both (behavior XYYX; right-most green) is harder compared to all other behaviors that use both stimuli (other green). Interestingly, this behavior (XYYX) is analogous to XOR and XNOR binary boolean functions, which are the only ones that are not linearly separable. All comparisons are statistically significant at the level of $P < 0.001$ and hold for all experimented bounding boxes.

%% file: sections/swarm.tex
\begin{figure}[t]
    \centering
    \includegraphics[width=0.5\textwidth]{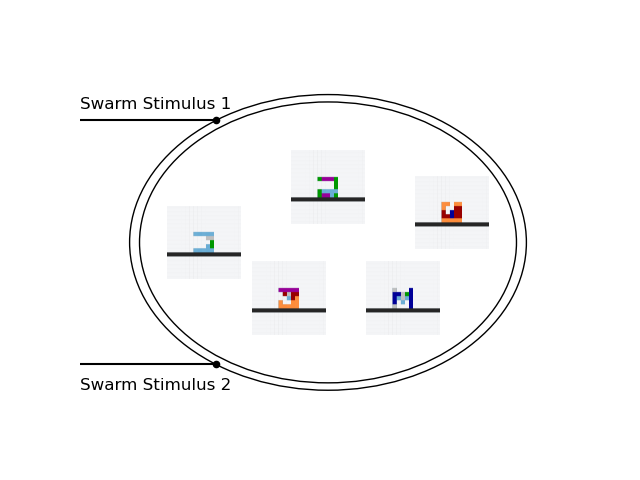}%
    \includegraphics[width=0.5\textwidth]{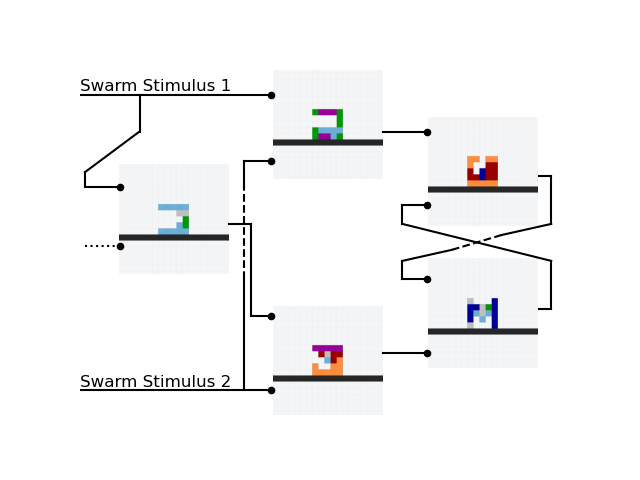}
    \\
    \vspace{-2em}
    \includegraphics[width=0.5\textwidth]{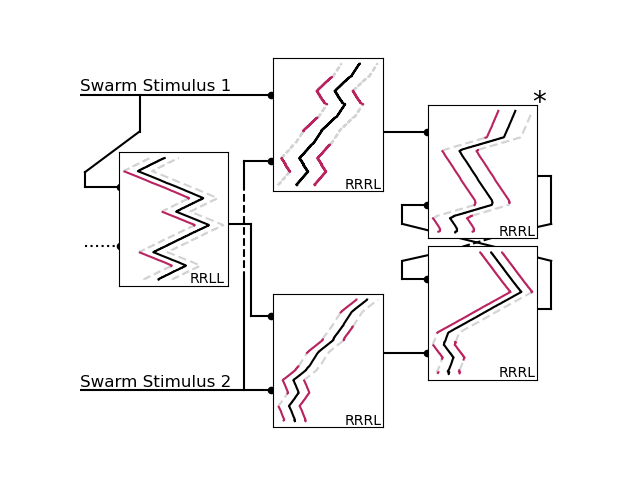}%
    \includegraphics[width=0.5\textwidth]{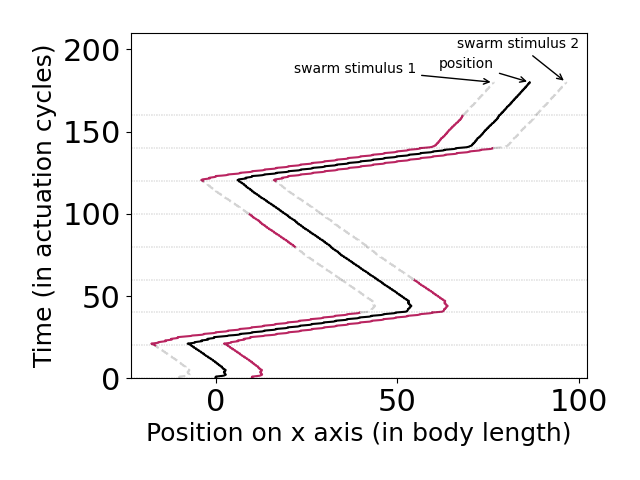}
    \caption{\textbf{Top row} High-level schematic of the swarm (left) and its inner wiring (right). Lines connecting to robots from the left represent stimuli and the ones on the right represent outputs. We assume that the behavior of one robot could determine the stimulus of another. \textbf{Bottom row} The behavior of each robot with respect to their immediate stimuli (left) and the behavior of the starred robot with respect to swarm stimuli (right). While each robot successfully performs the behavior that it evolved for with respect to its immediate stimuli, examination of the selected robot with respect to swarm stimuli shows complex behavior that would otherwise require memory to exhibit.}
    \label{fig:swarm-system}
\end{figure}

The above section demonstrated the evolution of robots exhibiting adaptive behavior in response to the two stimuli.  To the best of our knowledge, this is the first example of a closed-loop fully-morphological behavior in evolved robots.  The simplicity of the chosen behavior (more right or left) allows us to clearly see how shape change due to the stimuli-sensing materials affects the robots' behavior.  But this also invites a thought experiment: how could this proof of concept scale to create complex ``intelligent'' behaviors from this simple framework?  

In this section, we study how these robots can be grouped to create a swarm exhibiting more complex behavior. As a proof of concept, we use a subset of previously evolved robots and hand-design the swarm to show a certain behavior.

We start with an observation. Given the binary nature of the stimuli and the possible behaviors of the robot (moving in a positive or negative direction on the x-axis), the robots could be considered implementing logic gates from the perspective of an outside observer. Above, we evolved robots implementing all possible boolean functions by evolving an archive of robots exhibiting all possible behaviors (Fig.~\ref{fig:champions-behavior-fixed}). For instance, if we assume that a robot moving in the positive x direction on the x-axis is outputting a 1, robots exhibiting the RRRL behavior can be considered as implementing the NOT-AND boolean function. 

We can organize robots, by assuming that the behavior of one robot could determine the stimulus of another robot.  While this is arbitrary given our manually and abstractly designed stimuli, it's not a great stretch of the imagination to envision some analogous physical or chemical sensory apparatus that would allow voxels to sense the relative change in distance of a neighboring robot similar to how animals do (e.g. it became ``louder'' or its ``smell'' became stronger).

To demonstrate a more complex function derived from these morphological logic gates, we manually design a swarm implementing a D-type latch\footnote{While we hand-design the swarm as a proof of concept, one could evolve the design of the swarm towards achieving any target behavior as well.}. Fig.~\ref{fig:swarm-system} top left, shows the schematic of the swarm from an outside observer's perspective. The swarm consists of a single robot with behavior RRLL (leftmost robot, implements AND gate), and four robots with behavior RRRL (implement NAND gate). The swarm has two swarm stimuli, observed by three of the robots. The rest of the stimuli that robots observe are determined by others' behavior. The inner wiring of the swarm can be seen in Fig.~\ref{fig:swarm-system} top right, where wires on the left side of the robots represent stimuli and the wires on the right side of the robots represent the output of the robot (which determines the stimulus of another robot). If a robot's center of mass has moved in the positive x direction in the last actuation cycle it is considered to be outputting 1 and turns the stimulus on for another robot and vice versa. 


We plot the behavior of the swarm as a spacetime diagram in Fig.~\ref{fig:swarm-system} bottom row. The behavior of each individual and their immediate stimuli can be seen on the left. As expected, robots successfully perform the behaviors that are evolved for, seamlessly responding to changing stimuli patterns. However, if we observe the behavior of the starred robot (top-right robot) with respect to swarm stimuli (as opposed to the robot's immediate stimuli) in Fig.~\ref{fig:swarm-system} bottom right, we see that the robot is moving following the swarm stimulus 1, only when the swarm stimulus 2 is present, otherwise the robot keeps moving in the direction of last observed swarm stimulus 1. Given the right frame of reference (observing the robot's behavior with respect to the swarm stimuli) and certain assumptions (behavior of one robot determining the stimulus of another), we show that robots exhibit complex adaptive behavior, without any separately postulated brain, that would typically require memory to achieve. This achievement is, certainly, not a surprise, given the equivalence of robots' behavior with boolean functions. Once we have robots that can implement the NOT-AND function, we could build a swarm that would perform any computable behavior. We leave the design of such swarms as an exercise for the reader. 

%% file: sections/discussion.tex

The above results demonstrate ``intelligent'' morphological computation that adapts the robot's behavior via stimuli-educed shape change rather than by a dedicated control or brain-like mechanism. While we feel that this proof of concept is an important, and hopefully entertaining, demonstration, we fully acknowledge that the most complex and adaptive behaviors in natural systems are not fully morphological, but are an interplay of embodied cognition connecting the body and the brain.  

Recent literature had demonstrated the challenges of brain-body\ co optimization and noted fragile co-adaptation to be a major challenge of evolving these systems~\cite{cheney_difficulty_2016,mertan2024investigating}, further suggesting that increasing the adaptability and robustness of the brain and body to modification in the other could result in less fragile and thus more effective brain-body systems~\cite{cheney_scalable_2018,mertan2023modular}.  We hope that by creating more intelligent and adaptive bodies, this work helps to open doors to transfer additional computation to the morphology and enable less fragile co-adaptation.  


Lastly, we want to connect this research to recent ideas in artificial and natural evolution. Specifically, it has been argued that a key feature of natural evolution is that it works with material that is active. Cells that already have an array of functionalities come together to create multi-cellular structures, tissues, organs, and systems, creating a ``multiscale competence architecture'' where each scale exploits and regulates the already-existing functionalities of the scale below to achieve its goals~\cite{levin2023darwin}. This is considered to be an important feature of natural evolution and understanding/exploiting this in artificial evolution could have substantial effects~\cite{bongard_theres_2023,hartl_evolutionary_2024}. Inspired by these ideas, here we show a first pass at demonstrating that simple materials each with straightforward behaviors can be optimized in their emergent design to result in a complex and adaptive system. In future work, we hope to examine how we can consider such a system as a substrate for further evolution (potentially in combination with separately postulated brain models).

%% file: sections/conclusion.tex
We investigate ways in which complex, adaptive behavior can be created in virtual voxel-based soft robots without the use of a separately postulated brain. We hand-design simple muscle (that continuously expands and contracts) and sensory (that expands/shrinks in the presence or absence of a binary stimulus) voxel materials and show that evolution can create designs that exhibit closed-loop adaptive behavior, responding to two binary stimuli. Moreover, a swarm of such robots can exhibit even more complex behavior. In future work, we hope to study more ways of creating adaptive and complex behavior.

%% file: main.bbl
\begin{thebibliography}{10}
\providecommand{\url}[1]{\texttt{#1}}
\providecommand{\urlprefix}{URL }
\providecommand{\doi}[1]{https://doi.org/#1}

\bibitem{auerbach2010evolving}
Auerbach, J.E., Bongard, J.C.: Evolving cppns to grow three-dimensional physical structures. In: Proceedings of the 12th annual conference on Genetic and evolutionary computation. pp. 627--634 (2010)

\bibitem{beer1996toward}
Beer, R.D.: Toward the evolution of dynamical neural networks for minimally cognitive behavior. In: From Animals to Animats 4: Proceedings of the Fourth International Conference on Simulation of Adaptive Behavior. vol.~4, p.~421. MIT Press (1996)

\bibitem{bhatia2021evolution}
Bhatia, J., Jackson, H., Tian, Y., Xu, J., Matusik, W.: Evolution gym: A large-scale benchmark for evolving soft robots. Advances in Neural Information Processing Systems  \textbf{34},  2201--2214 (2021)

\bibitem{bongard2022er}
Bongard, J.: Evolutionary robotics lectures, \url{https://www.youtube.com/playlist?list=PLAuiGdPEdw0iyApypcLk_xBKjLchQM-Mg}

\bibitem{bongard2003evolving}
Bongard, J.C., Pfeifer, R.: Evolving complete agents using artificial ontogeny. In: Morpho-functional machines: the new species: designing embodied intelligence. pp. 237--258. Springer (2003)

\bibitem{bongard_theres_2023}
Bongard, J., Levin, M.: There’s {Plenty} of {Room} {Right} {Here}: {Biological} {Systems} as {Evolved}, {Overloaded}, {Multi}-{Scale} {Machines}. Biomimetics  \textbf{8}(1), ~110 (Mar 2023). \doi{10.3390/biomimetics8010110}

\bibitem{braitenberg1986vehicles}
Braitenberg, V.: Vehicles: Experiments in synthetic psychology. MIT press (1986)

\bibitem{cheney_difficulty_2016}
Cheney, N., Bongard, J., Sunspiral, V., Lipson, H.: On the {Difficulty} of {Co}-{Optimizing} {Morphology} and {Control} in {Evolved} {Virtual} {Creatures}. In: Proceedings of the {Artificial} {Life} {Conference} 2016. pp. 226--233. MIT Press, Cancun, Mexico (2016). \doi{10.7551/978-0-262-33936-0-ch042}

\bibitem{cheney2015evolving}
Cheney, N., Bongard, J., Lipson, H.: Evolving soft robots in tight spaces. In: Proceedings of the 2015 annual conference on Genetic and Evolutionary Computation. pp. 935--942 (2015)

\bibitem{cheney_scalable_2018}
Cheney, N., Bongard, J., SunSpiral, V., Lipson, H.: Scalable co-optimization of morphology and control in embodied machines. Journal of The Royal Society Interface  \textbf{15}(143),  20170937 (Jun 2018). \doi{10.1098/rsif.2017.0937}

\bibitem{cheney2013unshackling}
Cheney, N., MacCurdy, R., Clune, J., Lipson, H.: Unshackling evolution: evolving soft robots with multiple materials and a powerful generative encoding. In: Proceedings of the 15th annual conference on Genetic and evolutionary computation. pp. 167--174 (2013)

\bibitem{corucci_material_2016}
Corucci, F., Cheney, N., Lipson, H., Laschi, C., Bongard, J.: Material properties affect evolutions ability to exploit morphological computation in growing soft-bodied creatures. In: Proceedings of the {Artificial} {Life} {Conference} 2016. pp. 234--241. MIT Press, Cancun, Mexico (2016). \doi{10.7551/978-0-262-33936-0-ch043}

\bibitem{descartes_discourse_1993}
Descartes, 1596-1650, R.: Discourse on method ; and, {Meditations} on first philosophy. Third edition. Indianapolis : Hackett Pub. Co., [1993] ©1993 (1993), \url{https://search.library.wisc.edu/catalog/999718190702121}

\bibitem{eiben2015introduction}
Eiben, A.E., Smith, J.E.: Introduction to evolutionary computing. Springer (2015)

\bibitem{hartl_evolutionary_2024}
Hartl, B., Risi, S.L., Levin, M.: Evolutionary {Implications} of {Multi}-{Scale} {Intelligence} (Apr 2024). \doi{10.31219/osf.io/sp9kf}

\bibitem{hauser2011towards}
Hauser, H., Ijspeert, A.J., F{\"u}chslin, R.M., Pfeifer, R., Maass, W.: Towards a theoretical foundation for morphological computation with compliant bodies. Biological cybernetics  \textbf{105},  355--370 (2011)

\bibitem{hornby2001body}
Hornby, G.S., Pollack, J.B.: Body-brain co-evolution using l-systems as a generative encoding. In: Proceedings of the 3rd Annual Conference on Genetic and Evolutionary Computation. pp. 868--875 (2001)

\bibitem{joachimczak2016artificial}
Joachimczak, M., Suzuki, R., Arita, T.: Artificial metamorphosis: Evolutionary design of transforming, soft-bodied robots. Artificial life  \textbf{22}(3),  271--298 (2016)

\bibitem{joachimczak2012co}
Joachimczak, M., Wr{\'o}bel, B.: Co-evolution of morphology and control of soft-bodied multicellular animats. In: Proceedings of the 14th annual conference on Genetic and evolutionary computation. pp. 561--568 (2012)

\bibitem{karban2008plant}
Karban, R.: Plant behaviour and communication. Ecology letters  \textbf{11}(7),  727--739 (2008)

\bibitem{lakoff2008metaphors}
Lakoff, G., Johnson, M.: Metaphors we live by. University of Chicago press (2008)

\bibitem{lecun2015deep}
LeCun, Y., Bengio, Y., Hinton, G.: Deep learning. nature  \textbf{521}(7553),  436--444 (2015)

\bibitem{lehman2011evolving}
Lehman, J., Stanley, K.O.: Evolving a diversity of virtual creatures through novelty search and local competition. In: Proceedings of the 13th annual conference on Genetic and evolutionary computation. pp. 211--218 (2011)

\bibitem{levin2023darwin}
Levin, M.: Darwin’s agential materials: evolutionary implications of multiscale competency in developmental biology. Cellular and Molecular Life Sciences  \textbf{80}(6), ~142 (2023)

\bibitem{mace2023quality}
Mac{\'e}, V., Boige, R., Chalumeau, F., Pierrot, T., Richard, G., Perrin-Gilbert, N.: The quality-diversity transformer: Generating behavior-conditioned trajectories with decision transformers. In: Proceedings of the Genetic and Evolutionary Computation Conference. pp. 1221--1229 (2023)

\bibitem{mcculloch1943logical}
McCulloch, W.S., Pitts, W.: A logical calculus of the ideas immanent in nervous activity. The bulletin of mathematical biophysics  \textbf{5},  115--133 (1943)

\bibitem{mcgeer1990passive}
McGeer, T.: Passive dynamic walking. The international journal of robotics research  \textbf{9}(2),  62--82 (1990)

\bibitem{mertan2023modular}
Mertan, A., Cheney, N.: Modular controllers facilitate the co-optimization of morphology and control in soft robots. In: Proceedings of the Genetic and Evolutionary Computation Conference. pp. 174--183 (2023)

\bibitem{mertan2024investigating}
Mertan, A., Cheney, N.: Investigating premature convergence in co-optimization of morphology and control in evolved virtual soft robots. In: European Conference on Genetic Programming (Part of EvoStar). pp. 38--55. Springer (2024)

\bibitem{moravec1988mind}
Moravec, H.: Mind children: The future of robot and human intelligence. Harvard University Press (1988)

\bibitem{mouret_illuminating_2015}
Mouret, J.B., Clune, J.: Illuminating search spaces by mapping elites. arXiv:1504.04909 [cs, q-bio]  (Apr 2015), \url{http://arxiv.org/abs/1504.04909}

\bibitem{paul2006morphological}
Paul, C.: Morphological computation: A basis for the analysis of morphology and control requirements. Robotics and Autonomous Systems  \textbf{54}(8),  619--630 (2006)

\bibitem{pfeifer2006body}
Pfeifer, R., Bongard, J.: How the body shapes the way we think: a new view of intelligence (2006)

\bibitem{pfeifer2009morphological}
Pfeifer, R., G{\'o}mez, G.: Morphological computation--connecting brain, body, and environment. Creating brain-like intelligence: From basic principles to complex intelligent systems pp. 66--83 (2009)

\bibitem{pulvermuller2010active}
Pulverm{\"u}ller, F., Fadiga, L.: Active perception: sensorimotor circuits as a cortical basis for language. Nature reviews neuroscience  \textbf{11}(5),  351--360 (2010)

\bibitem{sims2023evolving}
Sims, K.: Evolving virtual creatures. In: Seminal Graphics Papers: Pushing the Boundaries, Volume 2, pp. 699--706 (2023)

\bibitem{stanley_compositional_2007}
Stanley, K.O.: Compositional pattern producing networks: {A} novel abstraction of development. Genetic Programming and Evolvable Machines  \textbf{8}(2),  131--162 (Jun 2007). \doi{10.1007/s10710-007-9028-8}

\bibitem{trewavas2003aspects}
Trewavas, A.: Aspects of plant intelligence. Annals of botany  \textbf{92}(1),  1--20 (2003)

\bibitem{wilson2002six}
Wilson, M.: Six views of embodied cognition. Psychonomic bulletin \& review  \textbf{9},  625--636 (2002)

\end{thebibliography}
